\documentclass[[final,3p,times,twocolumn]{elsarticle}
\makeatletter
\def\ps@pprintTitle{
	\let\@oddhead\@empty
	\let\@evenhead\@empty
	\def\@oddfoot{}%
	\let\@evenfoot\@oddfoot}
\makeatother
\usepackage{lineno}
\modulolinenumbers[5]

\journal{Journal of Image and Vision Computing}
\usepackage{framed,multirow}
\usepackage{amssymb}
\usepackage{latexsym}
\usepackage{url}
\usepackage{xcolor}
\usepackage{xspace}                      
\definecolor{newcolor}{rgb}{.8,.349,.1}
\usepackage{mathtools}
\usepackage{algorithm}
\usepackage{graphics}

\newcommand{\ie}{\textit{i.e.~}{\protect\nopagebreak[4]}}

\newcommand{\etal}{{\em{et~al.}}\xspace}

\newtheorem{theorem}{Theorem}[section]

\newtheorem{definition}[theorem]{Definition}
\newenvironment{tequation}{\small\begin{equation}}{\end{equation}}
\def\Vec#1{{\boldsymbol{#1}}}
\def\Mat#1{{\boldsymbol{#1}}}
\DeclareMathOperator*{\argmin}{arg\,min}

\newcommand{\capspace}{{\vspace{-0.12in}}} 
\newcommand{\picspace}{{\vspace{-0.2in}}} 

\begin{document}
	
	\begin{frontmatter}


\title{Convex Class Model on Symmetric Positive Definite Manifolds}
\author{Kun {Zhao}\corref{cor1}} 
\cortext[cor1]{Corresponding author.} 
\ead{kun.zhao@uq.net.au}
\author{Arnold {Wiliem}}
\author{Shaokang {Chen}}
\author {Brian C. {Lovell}}
\address{School of Information Technology and Electrical Engineering, \\
	The University of Queensland, St Lucia, Brisbane, QLD 4072, Australia}

\begin{abstract} 
The effectiveness of Symmetric Positive Definite (SPD) manifold features has been proven in various computer vision tasks.
However, due to the non-Euclidean geometry of these features, existing Euclidean machineries cannot be directly used.
In this paper, we tackle the classification tasks with limited training data on SPD manifolds.
Our proposed framework, named Manifold Convex Class Model, represents each class on SPD manifolds using a convex model, and classification can be performed by computing distances to the convex models.
We provide three methods based on different metrics to address the optimization problem of the smallest distance of a point to the convex model on SPD manifold.
The efficacy of our proposed framework is demonstrated both on synthetic data and several computer vision tasks including object recognition, texture classification, person re-identification and traffic scene classification.
\end{abstract}
\end{frontmatter}

\section{Introduction}
\label{intro}
Despite of the recent fast progress in computer vision, most of effective machine learning paradigms demand large numbers of labeled samples to obtain a high performance.
However, collecting a large number of labeled samples is often difficult, time-consuming and expensive. For instance, obtaining large reliable and balanced data in medical imaging is quite challenging~\cite{kalunga2015data}. 
Limited training data may fail to represent the true distribution of the data and can cause severe problems such as overfitting.    
In this paper, we tackle the classification tasks with limited training data  where the data are in the forms of Symmetric Positive Definite (SPD) matrices.
In various applications, data is
often represented in terms of SPD matrices. 
For example, the Diffusion Tensor Imaging (DTI) represents each voxel by a SPD matrix~\cite{pennec2006riemannian,arsigny2007geometric}. Tuzel \etal~\cite{tuzel2008pedestrian} introduced SPD features for texture matching and classification, where region covariance matrix were used as one image descriptor.
Region covariance matrices were believed to be discriminative as the second-order statistics of regional features are captured ~\cite{tuzel2008pedestrian,tosato2010multi}. 
In the context of image set classification or video-based classification, representing image sets or videos via
set-based covariance also has proven to be effective in several computer vision tasks~\cite{huang2015log,huang2017geometry}. As the second-order statistic of a set of samples, the covariance matrix encoded the feature correlations specific to each class, which leads to discriminate image sets/videos of different classes~\cite{huang2015log,huang2017geometry}.
If Euclidean geometry is utilized on these
SPD matrices, it may cause unacceptable results such as the swelling effect which occurred in DTI~\cite{arsigny2007geometric}.
To overcome this problem, Pennec~\etal\cite{pennec2006riemannian} proposed to endow a Riemannian manifold on these SPD matrices, which preserves the non-Euclidean structures originated from these data. 
It is also noteworthy that although manifold features are robust to certain variations, they still do not address all possible variations in image domain, such as white noise~\cite{yin2016kernel}. 

Inspired by the success of nearest convex models for classification tasks with limited data in Euclidean space~\cite{cevikalp2009large,takahashi2011construction,chen2014matching,chen2013improved,cevikalp2008nearest},  we propose a novel framework on SPD manifolds, here called Manifold Convex Class Model (MCCM), which serves as a generalization of convex model from Euclidean space.
More specifically,  the proposed MCCM represents each class using a convex class model which includes all convex combinations of the data points within the class.
Classification is then performed by computing distances to the convex class models.

\begin{figure}
	\centering
	\includegraphics[width=0.4\textwidth,keepaspectratio]{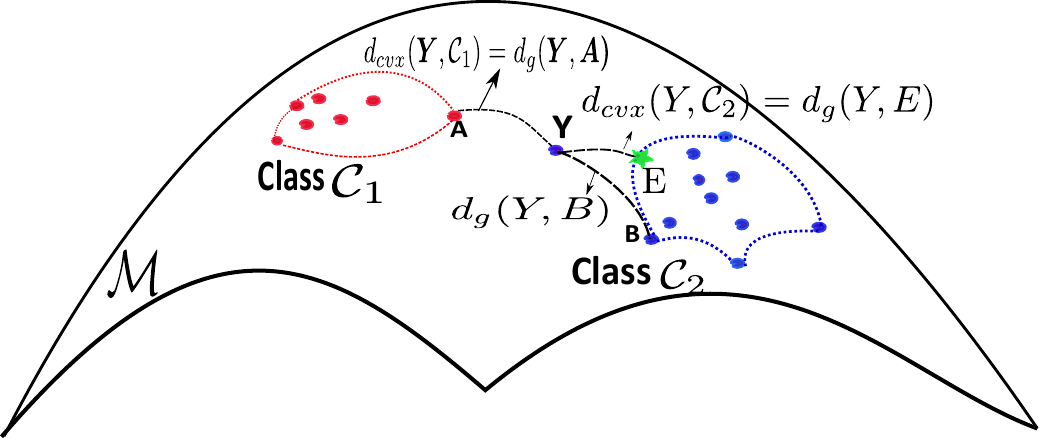} 
	\caption{ Conceptual illustration of the proposed Manifold Convex Class Model (MCCM).
		Given two classes of SPD points, $\mathcal{C}_1$ and $\mathcal{C}_2$, for each class, we construct a convex model by the convex combinations of the points within the class under certain conditions.
		We classify the SPD point $\Mat{Y}$ to class $\mathcal{C}_2$, as the distance between the convex class model of class $\mathcal{C}_2$ and the point $\Mat{Y}$~(refer to Eqn.~\eqref{eq:model}) is smaller than that of class $\mathcal{C}_1$ . 
	}                            
	\label{fig:illlustration}\picspace
	
\end{figure}
Figure~\ref{fig:illlustration} illustrates the concept of our proposed Manifold Convex Class Model (MCCM). We consider a convex model to represent each class.
To classify a SPD point $\Mat{Y}$, we compute the smallest distance between the convex class model and the point $\Mat{Y}$. As shown in Figure~\ref{fig:illlustration}, $d_{cvx}(\Mat{Y},\mathcal{C}_1)=d_g(\Mat{Y},\Mat{{A}})$, $d_{cvx}(\Mat{Y},\mathcal{C}_2)=d_g(\Mat{Y},\Mat{E})$. The generated point $\Mat{E}$ is produced from a convex combination of points in $\mathcal{C}_2$, and $\operatorname{d}_{g}$ is the geodesic distance defined in Eqn.~\eqref{Eqn:dist_SPD}.
MCCM classifies $\Mat{Y}$ to  $\mathcal{C}_2$ as the convex model distance $d_{cvx}(\Mat{Y},\mathcal{C}_2)<d_{cvx}(\Mat{Y},\mathcal{C}_1)$.
However, the nearest neighbor approach incorrectly classifies $\Mat{Y}$ to $\mathcal{C}_1$ as it finds the nearest point $\Mat{A}$ in $\mathcal{C}_1$, $\Mat{B}$ in $\mathcal{C}_2$ and the geodesic distance $\operatorname{d}_{g}(Y,A)<\operatorname{d}_{g}(Y,B)$.

Unfortunately, due to
the non-Euclidean structure on manifolds, 
it is not straightforward to construct a convex class model.
This paper starts by investigating the existence of convex model on SPD manifolds. Then, we formulate the classification problem based on the convex model.   
More specifically, the contributions of this work are listed as follows:  
\begin{itemize}
	\item Proposing a novel mathematical framework, here called Manifold Convex Class Model~(MCCM), to solve classification tasks with limited training data where the data are in SPD matrices; especially, we define the convex combinations
	and the nearest distance to convex model over the SPD manifolds;
	\item Proposing a solution to the optimization problem of the framework, MCCM-FM, which preserves the intrinsic manifold structure by the use of Affine Invariant Riemannian Metric~(AIRM);
	to reduce the computational complexity, two approximate solutions are provided: MCCM-CS and MCCM-LE;
	\item The experimental results show that MCCM significantly outperforms other intrinsic classifiers and has competitive performance to recent methods on several computer vision applications: object recognition, texture classification, person re-identification and traffic scene classification;
	\item Providing additional analysis on the approximation errors and running time of the three proposed solutions; 
	\item Compared with the Geodesic Nearest Neighbor(Geo-NN) in terms of the sensitivity to noise and limited training data, the proposed framework, especially MCCM-FM, is more robust to image noise and less sensitive to the number of training data.  
\end{itemize}

We presented a preliminary version of the classification approach based on convex model for SPD points in~\cite{zhao2016convexhull}. We note that the work discussed here differs from the one in~\cite{zhao2016convexhull} in three aspects. First, in this work, we explore and study the convex model topic from the manifold intrinsic geometry. This leads to a solution to solve the convex model optimization problem, which seeks to preserve the intrinsic structure as much as possible. In our previous work, we only presented one approximate solution.
Second,  
in this work, we present alternative solutions to the optimization of convex model on SPD manifolds, which are more efficient and effective. 
Third, in the experimental section, our previous work only reported results on three applications: object recognition, texture classification and person re-identification. In this paper, we also explore one additional application: traffic scene classification.

We continue this paper as follows. 
Section~\ref{SPD} briefly describes the geometry of SPD manifolds.
Section~\ref{re_work} reviews the related work.
Section~\ref{SPD_CH} presents the proposed MCCM methods over SPD manifolds. 
Section~\ref{ex_results} shows the experimental results for four computer vision applications.
Section~\ref{further} provides further analysis on the proposed MCCM methods.
Section~\ref{conclusion} summarizes the main findings and presents the potential future work.

\section{ Symmetric Positive Definite Manifold}
\label{SPD}
The Symmetric Positive Definite~(SPD) manifold is one specific type of Riemannian manifold.
It is a smooth manifold where the tangent space is endowed with a Riemannian metric~\cite{pennec2006riemannian}.
The Riemannian metric allows us to define various geometric notions such as the geodesic distance.
One of the most popular Riemannian metrics for SPD manifolds is the Affine Invariant Riemannian Metric~(AIRM)~\cite{pennec2006riemannian} defined as:
\begin{tequation}
	\label{AIRM}
	\left\langle\Vec{x},\Vec{z} \right\rangle_{\Mat{Y}}=\operatorname{Tr}\left(\Mat{Y}^{-1}\Vec{x} \Mat{Y}^{-1}\Vec{z}\right) \textrm{ ,} 
\end{tequation}\noindent
where $\Vec{x}$ and $\Vec{z}$ are tangent vectors (symmetric matrices, not necessarily definite nor positive) on the tangent space at point $\Mat{Y}$, and $\operatorname{Tr}(\centerdot)$ is the matrix trace.
A point $\Vec{z}$ on the tangent space at $\Mat{Y}$ can be mapped to the SPD manifold by the following manifold exponential function~\cite{pennec2006riemannian}:

\begin{tequation}
	\label{exp_map_y}
	\operatorname{exp}_{\Mat{Y}}(\Vec{z})=\Mat{Y}^{\frac{1}{2}}\operatorname{exp}(\Mat{Y}^{-\frac{1}{2}}\Mat{z}\Mat{Y}^{-\frac{1}{2}})\Mat{Y}^{\frac{1}{2}}\textrm{ ,}
\end{tequation}\noindent
where $\operatorname{exp}(\centerdot)$ denotes the matrix exponential.
The inverse function of Eqn.~\eqref{exp_map_y} maps a
point $\Mat{Z}$ on the SPD manifold to the tangent space at  $\Mat{Y}$:
\begin{tequation}
	\label{log_map_y}
	\operatorname{log}_{\Mat{Y}}(\Mat{Z})=\Mat{Y}^{\frac{1}{2}}\operatorname{log}(\Mat{Y}^{-\frac{1}{2}}\Mat{Z}\Mat{Y}^{-\frac{1}{2}})\Mat{Y}^{\frac{1}{2}}\textrm{ ,}
\end{tequation}\noindent
where $\operatorname{log}(\centerdot)$ denotes the matrix logarithm.

The geodesic distance
between two SPD points is defined as the length of the shortest
curve on the manifold. A widely used distance function is defined using the Affine Invariant Riemannian Metric~(AIRM)~\cite{pennec2006riemannian}:

\begin{tequation}
	\label{Eqn:dist_SPD}
	\operatorname{d}_{g}(\Mat{X},\Mat{Y})=||\operatorname{log}_{\Mat{X}}(\Mat{Y})||_{\Mat{X}}=\sqrt{\operatorname{Tr}(\operatorname{log}^2(\Mat{X}^{- \frac{1}{2}}\Mat{Y} \Mat{X}^{- \frac{1}{2}}))}
	\textrm{ ,}
\end{tequation}\noindent
\noindent where $\operatorname{Tr}$ is the matrix trace computation.
The AIRM possesses some useful properties such as invariance to affine transformations of the input matrices.

Another widely used distance function for SPD points is derived from the Log-Euclidean metric~\cite{arsigny2007geometric}:

\begin{tequation}
	\label{Eqn:dist_LED}
	\operatorname{d}_{LE}(\Mat{X},\Mat{Y})=||\operatorname{log}\Mat{X}-\operatorname{log}\Mat{Y}||_F\textrm{ .} 
\end{tequation}\noindent

Compared to the distance $\operatorname{d}_{g}$  defined in Eqn.~\eqref{Eqn:dist_SPD}, the computational cost of $\operatorname{d}_{LE}$ defined in Eqn.~\eqref{Eqn:dist_LED} is often less due to the use of Euclidean computations in the matrix logarithm domain. 

\section{Related Work}
\label{re_work}

In the recent literature, to solve classification problems for SPD points, kernelized learning methods are commonly used~\cite{harandi2015extrinsic,jayasumanaetal2013,vemulapalliPillai2013}.
Nevertheless, choosing an appropriate kernel has been shown to be non-trivial for manifolds and a poor choice of kernel might lead to inferior performance~\cite{vemulapalliPillai2013,feragen2015geodesic}.
Feragen~\etal demonstrated that most of the kernels applied on Riemannian manifolds either ignored the intrinsic structure or the positive definiteness constraint~\cite{feragen2015geodesic}.
An alternative method to classify SPD points is to map all the manifold points onto a tangent space at a designated location~\citep{tuzel2008pedestrian,harandi2015extrinsic}. 
Once mapped, any Euclidean-based classifiers can be employed.
Unfortunately, this mapping may adversely affect the performance
since it will significantly distort the manifold structure in
regions far from the origin of the tangent space~\citep{jayasumanaetal2013}. 

Therefore, developing classification methods that incorporate the intrinsic structures on manifolds is critical to process these data appropriately. 
In its simplest form, one could use Geodesic Nearest Neighbor that utilized the geodesic distance along the manifolds~\cite{pennec2006riemannian}. It has
been shown that this approach can achieve good accuracy. However,
it is generally sub-optimal when the training data is scarce and do not suffice to cover the class regions.  
Recently, several sparse coding methods on SPD manifolds were proposed in the literature.
Sivalingam~\etal\cite{sivalingam2010tensor} proposed tensor sparse coding for SPD manifolds. Since their method used log-determinant divergence to model the loss function, it suffers from high computational complexity.  Also, the manifold structure is not well-preserved by the log-determinant divergence, thus the accuracy is not optimal. 
Ho~\etal~\cite{xie2013nonlinear} proposed a nonlinear generalization of dictionary learning and sparse coding on SPD manifolds using the geodesic distances. 
Similarly, Cherian~\etal proposed Riemannian sparse coding
for SPD manifolds~\cite{cherian2014riemannian}, which minimizes the geodesic distance to the matrix addition of weighted SPD points.
However, as shown in our experiments, sparse coding and dictionary learning on SPD manifolds is usually not effective when the training set is relatively small. 

To solve the classification problem with limited data on SPD manifold, we intend to generalize the convex model developed in the Euclidean counterpart. 
In Euclidean space where data is represented by vectors, convex models such as affine hull~\cite{vincent2001k}, convex
hull~\cite{qing2008nearest,zhou2009nearest}, bounding hypersphere~\cite{tax2004support} and bounding hyperdisk~\cite{cevikalp2008nearest}, are used to tackle the limited data problem.
For instance, given a set of points $\mathcal{S} = \{
\Vec{x}_i \}_{i=1}^N$, $\Vec{x}_i \in \mathbb{R}^d$, the convex hull of $\mathcal{S}$ is the intersection of all half-spaces which contain $\mathcal{S}$. 
Alternatively speaking, the convex hull  $\mathcal{C}$, generated from the set $\mathcal{S}$, includes all possible convex combination of the members in $\mathcal{S}$:
\begin{tequation}
	\label{eq:c_eu}
	\mathcal{C} = \left\{ \sum_{i=1}^N w_i \Vec{x}_i \right\}\textrm{, }  \sum_{i=1}^N w_i = 1\textrm{, } w_i \geq 0\textrm{, } i = 1,2,...,N \textrm{ .}
\end{tequation}\noindent
The convex hull distance of a point $\Vec{y}$ to  $\mathcal{C}$
is defined as the minimal distance of $\Vec{y}$ to any point in the
convex hull $\mathcal{C}$ by the
following:
\begin{tequation}
	\setlength{\abovedisplayskip}{2pt}
	\setlength{\belowdisplayskip}{3pt}
	\centering
	\label{eq:cvx_eu}
	\begin{split} 
		& \operatorname{d}_{cvx}^2(\Vec{y},\mathcal{C})  =  \min  ||\Vec{y}-\tilde{\Vec{x}}||_{2}^{2}  =  \min ||\Vec{y} -\sum_{i=1}^{N}w_i \Vec{x}_i||{_{2}^{2}}\textrm{ ,}\\
		& s.t. \textrm{ } \sum\limits_{i = 1}^N {{w_i}}  = 1\textrm{, } {w_i} \ge
		0\textrm{, }i = 1,2,...,N\textrm{ .}
	\end{split}
\end{tequation}\noindent
The convex models have been used in numerous different applications, for instance, image registration, handwritten
shape detection and Glaucoma detection~\cite{yang1999image}.
In the computer vision community, convex model is also widely used to address image classification problems, such as face recognition and object recognition~\cite{tuzel2008pedestrian,cevikalp2010face}.
Thus, the similar benefits could be obtained if it is applied to the more complex data represented using SPD matrices.
To that end, in our work, we generalize the convex model from Euclidean counterpart to SPD manifolds.

\section{Manifold Convex Class Model (MCCM) for SPD Manifold}
\label{SPD_CH} 
The success of convex model in Euclidean space inspires us to extend this model to solve the classification problem over SPD manifold, especially when the training points are limited.
In the following part of this section,
we first introduce the definitions related with Manifold Convex Class Model (MCCM). Then we further provide three practical solutions for the optimization of the nearest convex model classification on SPD manifolds.
\subsection{Definitions for Manifold Convex Class Model}
\begin{figure}[!t]
	\centering
	\includegraphics[width=0.4\textwidth,keepaspectratio]{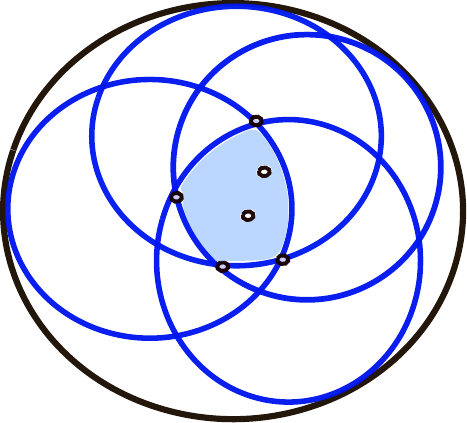} \capspace     \caption
	{
		An illustration of a ball hull constructed by the intersection
		of horoballs~\cite{fletcher2011horoball}.
	}
	\label{fig: ballhull}\picspace
\end{figure}

In Euclidean Space,  one way to construct a convex model of a set of points $\mathcal{S}$ is to use the intersection of all half-spaces that contain $\mathcal{S}$. 
However, it is not trivial to generalize this convex model construction formulation to the SPD manifold due to the space curvature. SPD manifolds generally do not admit half-spaces~\cite{fletcher2011horoball}. Therefore, one may wonder whether a convex model exists and can be formulated in the SPD manifold space.
To that end, in~\cite{fletcher2011horoball}, Fletcher~\etal used the horoball as the replacement of half-space on the SPD manifolds.
The horoball is a ball fixed at a point, whose radius is allowed to grow to infinity.
In Euclidean space, a horoball generates a half-space passing through a fixed point.
In SPD manifold space, a horoball is not flat due to the curvature of this space~\cite{fletcher2011horoball}.
However, it is guaranteed to be convex and closed, acting as an effective proxy for the half-space to define the construction of a convex model in SPD manifolds. 
More specifically,
given a group of points $\mathcal{S}$ on the SPD manifold, $\mathcal{S} = \{ \Mat{X}_i \}_{i=1}^N$, 
the intersection of all horoballs containing every single member of $\mathcal{S}$ would yield a ball hull. The ball hull is a convex model containing all the points in $\mathcal{S}$ on SPD manifolds~\cite{fletcher2011horoball}.
Figure~\ref{fig: ballhull} illustrates how the horoballs can be used to construct a ball hull.
This horoball concept is reformulated on SPD manifolds using a convex function called the Busemann function~\cite{fletcher2011horoball}. We refer readers to~\cite{fletcher2011horoball} for further details.

As a convex model exists in SPD manifolds, the next step is to study on how to use this concept for our problems.
To use the convex model as the tool to perform the classification tasks,
there are two questions to be
addressed: (1) Which is the most suitable convex model formulation on SPD manifold for our purpose, 
and (2) for a query point $\Mat{Y}$, how to compute the distance between $\Mat{Y}$ and a convex model?

Inspired by the Euclidean counterpart, the most convenient way to define a convex model of the set of SPD points is to use all the convex combinations of the point set. Here, we adapt this formulation into SPD manifold as follows.
\begin{definition}
	The convex model of a group of SPD points $\mathcal{S} = \{ \Mat{X}_i \}_{i=1}^N$,  is a convex set that contains all the convex combinations of these points and is formulated as:
	\begin{tequation}
		\label{eq:cv_spd}
		\begin{split}
			\mathcal{C}_{SPD} =  \bigoplus_{i=1}^N  w_i  \Mat{X}_i \textrm{, } 
			s.t. \textrm{ }  \sum\limits_{i = 1}^N {{w_i}}  = 1\textrm{, } {w_i} \ge
			0\textrm{, }i = 1,2,...,N\textrm{ ,}
		\end{split}
	\end{tequation}\noindent where
	$w_i\in \mathbb{R}_{0}^{+}$ and $\bigoplus$ is the convex combination operator over the SPD manifold. 
\end{definition}

The definitions and properties of convex combination operator in a metric space can be found in Ter{\'a}n and Molchanov's work~\cite{teran2006law}. 
Moreover, Ginestet~\etal\cite{ginestet2012weighted} proposed to use the weighted Fr{\'e}chet mean, which minimizes a weighted sum of squared distances, as a convex combination
operator in a metric space. They further showed that the Fr{\'e}chet mean operator actually allows the construction of convex models in a metric space~\cite{ginestet2012weighted}.
Following this intuition, in the metric space specified by a SPD manifold, we can construct the convex model of the given points by computing Fr{\'e}chet means.  
Thus, the convex model on a group of SPD points then can be illustrated using the following definition:
\begin{definition}
	The weighted sum of squared distances allows the construction of a convex model  $\mathcal{C}_{SPD}$ for the group of SPD points, $\mathcal{S} = \{ \Mat{X}_i \}_{i=1}^N$:
	\begin{tequation}
		\label{eq:cv_spd1}
		\begin{split}
			\mathcal{C}_{SPD} =\left\{   \tilde{\Mat{X}}\vert \forall\tilde{\Mat{X}} \in\argmin_{\tilde{\Mat{X}}} \sum_{i = 1}^N w_i {\operatorname{d_g}} ^2( \Mat{X}_i, \tilde{\Mat{X}} ) \textrm{, }
			s.t.\sum_{i=1}^N w_i = 1\textrm{, } w_i \geq 0 \right\}\textrm{ ,}
		\end{split}
	\end{tequation}\noindent where $ \operatorname{d_g}$ is the geodesic distance and $w_i\in \mathbb{R}_{0}^{+}$  is the
	weight of the $i$-th point $\Mat{X}_i$. 
\end{definition}
\noindent
This definition shows that the convex model of $\mathcal{S}$ is the set of all $\tilde{\Mat{X}}$ that minimized the weighted sum of squared distances to each point $\Mat{X}_i$ within $\mathcal{S}$.
Note that unlike the other Riemannian manifolds, the weighted mean solution in SPD manifolds always exists and is unique~\cite{moakher2005differential}. 

After we present the mathematical definition of a convex model in SPD manifolds, we then need to define how to compute the distance between a query point $\Mat{Y}$ and the convex model $\mathcal{C}_{SPD}$.
According to Udriste~\etal~\cite{udriste1994convex}, the distance between a query point $\Mat{Y}$ and a convex model $\mathcal{C}_{SPD}$ can be defined as the smallest geodesic distance from a point to a convex model on the manifold: 
\begin{tequation} 
	\label{eq:model}
	\operatorname{d_{cvx}}^2(\Mat{Y},  \mathcal{C}_{SPD} )= 
	\min{\operatorname{d_g}}^2(\Mat{Y}\textrm{, }   \tilde{\Mat{X}}), \forall\tilde{\Mat{X}}\in\mathcal{C}_{SPD}\textrm{ .} 
\end{tequation}
\noindent
Although Udriste~\etal~\cite{udriste1994convex} describes this, it is still not clear how to address this optimization problem. In our work, we will describe several ways to address this.
\subsection{Classification Based on Manifold Convex Class Model}
Assuming that we have the solution for Eqn.~\eqref{eq:model}, we are ready to define  the classifier based on the nearest convex model on SPD manifolds. Similar to Euclidean space, given a $m$-class classification problem, the classifier based on the nearest convex model on SPD manifolds can be formulated as:
\begin{tequation} 
	\label{eq:CH_Classifier}
	F(i)=\argmin_{i}\operatorname{d}_{cvx}^2
	(\Mat{Y}, \mathcal{C}_{SPD}^{i})\textrm{ ,} 
\end{tequation}\noindent
where $\mathcal{C}_{SPD}^{i}$ defined in Eqn.~\eqref{eq:cv_spd1} is the convex class model for the training class/category $i$ ($i={1,...,m}$, where $m$ is the number of classes/categories).   
To classify a query point $\Mat{Y}$,
one needs to compute the distance between $\Mat{Y}$ and each convex class model $\mathcal{C}_{SPD}^{i}$.
Finally, the query point $\Mat{Y}$ would be assigned the label of the training class wherein the geodesic distance between the convex class model $\mathcal{C}_{SPD}^{i}$ and $\Mat{Y}$ is the smallest.

It is important to solve the optimization problem presented in Eqn.~\eqref{eq:model}.
In the following sections, we describe three solutions to address this. More specifically, we first present a solution that seeks to preserve the intrinsic structure as much as possible.
Then, to speed up the computation, we propose two approximate solutions based on different relaxation conditions.
\subsection{Manifold Convex Class Model Based on Fr{\'e}chet Mean --- MCCM-FM}\label{SPD-intr}

\begin{figure}[!t]
	\centering
	\includegraphics[width=0.5\textwidth,keepaspectratio] {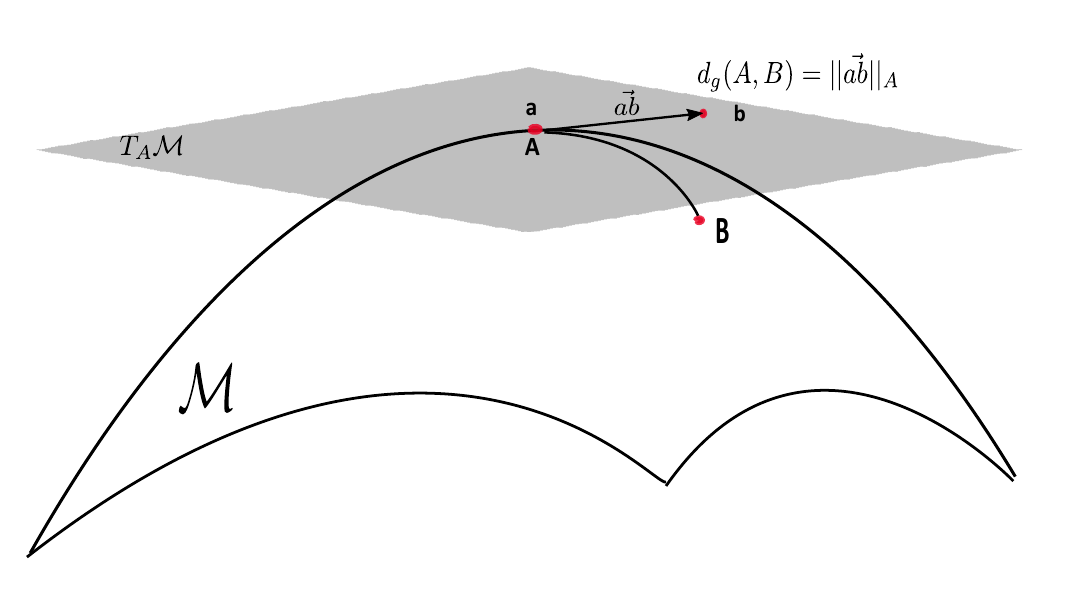} 
	\caption
	{Geodesic distance between SPD points.
	}
	\label{fig: GEO_dist}
\end{figure} 
To solve the optimization problem in Eqn.~\eqref{eq:CH_Classifier}, one needs to 
compute the distances between the query point $\Mat{Y}$ and each convex class model in the training set.
As shown in Eqn.~\eqref{eq:model}, this can be carried out by
determining the point $\tilde{\Mat{X}}$ in the convex class model, which has the minimum distance to the query point $\Mat{Y}$.
The point $\tilde{\Mat{X}}$ is one member of the convex model which is described by Eqn.~\eqref{eq:cv_spd1}.
When the weights $\{ w_i \} _{i=1}^{N} $ are fixed, $\tilde{\Mat{X}}$ becomes the weighted Fr{\'e}chet mean 
solution of the points in $\mathcal{C}_{SPD}$~\cite{moakher2005differential}.
Thus, convex model defined by Eqn.~\eqref{eq:cv_spd1} contains all the possible weighted Fr{\'e}chet means which are generated by varying the weights $\left\lbrace w_i\right\rbrace_{i=1}^{N}$ under the constraint of weights summation equals one.
By using Eqn.~\eqref{eq:cv_spd1} we can rewrite Eqn.~\eqref{eq:model} as follows:
\begin{small}
	\setlength{\abovedisplayskip}{3pt}
	\setlength{\belowdisplayskip}{3pt}
	\begin{align}\label{Eqn:cv_spd_mach1}
		\setlength{\abovedisplayskip}{2pt}
		\setlength{\belowdisplayskip}{2pt}	
		& \operatorname{d_{cvx}}^2(\Mat{Y},  \mathcal{C}_{SPD} )= \min \operatorname{d}_g^2 (\Mat{Y}, \Mat{\tilde{X}})\textrm{, }
		\nonumber  \\
		& s.t.\textrm{  }  \Mat{\tilde{X}} = \argmin_{\Mat{\tilde{X}}} \sum_{i=1}^N w_i \operatorname{d}_g^2 ( \Mat{\tilde{X}}, \Mat{X}_i )\textrm{, }
	\sum_{i=1}^N w_i = 1\textrm{, }  w_i \geq 0\textrm{ .}
	\end{align}
\end{small}\noindent Note that due to the weighted Fr{\'e}chet mean formulation, the above problem requires us to solve another optimization problem in the regularization part.
This makes solving the above problem extremely challenging.
To this end, we first examine how the geodesic distance $\operatorname{d}_g$ is derived. 
As described by Pennec~\etal~\cite{pennec2006riemannian}, the geodesic distance between $\Mat{A}$ and $\Mat{B}$ can be determined by calculating
the magnitude of the vector $\vec{\Mat{{a}}\Mat{b}}$ in the tangent space at $\Mat{A}$, where $\Mat{a} = \operatorname{log}_{\Mat{A}} (\Mat{A})$ and $\Mat{b} = \operatorname{log}_{\Mat{A}} (\Mat{B})$ are the projection of both $\Mat{A}$ and $\Mat{B}$ onto the tangent space at $\Mat{A}$, respectively. 
The vector magnitude can be calculated by using the given Riemannian metric (\ie the AIRM).
Figure~\ref{fig: GEO_dist} illustrates this relationship.
It is noteworthy to mention that the distance of all geodesic passing through $\Mat{A}$ can be determined by calculating the magnitude of vectors in the tangent space at $\Mat{A}$~\cite{pennec2006riemannian}.

We will exploit this fact to address Eqn.~\eqref{Eqn:cv_spd_mach1}. 
The aim of our framework is to find the smallest geodesic distance from $\Mat{Y}$ to the points in the convex model. 
Thus, preserving the geodesic distances 
between $\Mat{Y}$ and all points in the convex model are critical. 
To that end,
we solve the problem in the tangent space at $\Mat{Y}$, where the aforementioned geodesic distances are fully preserved.
Note that, in the tangent space at $\Mat{Y}$,  $\Mat{Y}$ becomes the origin $0$.
Thus, Eqn.\eqref{Eqn:cv_spd_mach1} can be rewritten as follows:
\begin{tequation}
	\setlength{\abovedisplayskip}{3pt}
	\setlength{\belowdisplayskip}{3pt}
	\label{Eqn:cv_spd_mach1_logy_2}
	\begin{split}
		\operatorname{d_{cvx}}^2(\Mat{Y},  \mathcal{C}_{SPD})=\min||\sum_{i=1}^N w_i\operatorname{log}_{\Mat{Y}}(\Mat{X}_i)||{^2_{\Mat{Y}}}
		\textrm{, }
		& s.t.\textrm{  } 
		\sum_{i=1}^N w_i = 1,\, w_i \geq 0\textrm{ .} 
	\end{split}
\end{tequation}\noindent
Finally, we substitute the AIRM in Eqn.~\eqref{AIRM} and Eqn.~\eqref{log_map_y} into Eqn.~\eqref{Eqn:cv_spd_mach1_logy_2}. We have the following optimization to solve:

\begin{tequation}
	\setlength{\abovedisplayskip}{3pt}
	\setlength{\belowdisplayskip}{3pt}
	\label{eq:intr_model}
	\begin{split}
		&\argmin_{ \{w_i\}_{i=1}^N}||\sum_{i=1}^N w_i\operatorname{log}_{\Mat{Y}}(\Mat{X}_i)||^2_{\Mat{Y}}\\ 
		&=\argmin_{ \{w_i\}_{i=1}^N} \operatorname{Tr}(\Mat{Y}^{-1}\sum_{i=1}^{N}(w_{i}\Mat{Y}^{\frac{1}{2}}\Mat{L}_{i}\Mat{Y}^{\frac{1}{2}})\Mat{Y}^{-1}\sum_{i=1}^{N}(w_{i}\Mat{Y}^{\frac{1}{2}}\Mat{L}_{i}\Mat{Y}^{\frac{1}{2}}))\textrm{ ,} \\
		&=\argmin_{ \{w_i\}_{i=1}^N} \operatorname{Tr}(\Mat{Y}^{-1}\Mat{Y}^{\frac{1}{2}}\sum_{i=1}^{N}(w_{i}\Mat{L}_{i})\Mat{Y}^{\frac{1}{2}}\Mat{Y}^{-1}\Mat{Y}^{\frac{1}{2}}\sum_{i=1}^{N}(w_{i}\Mat{L}_{i})\Mat{Y}^{\frac{1}{2}})\textrm{ ,} \\
		&=\argmin_{ \{w_i\}_{i=1}^N} \operatorname{Tr}(\Mat{Y}^{\frac{1}{2}}\Mat{Y}^{-1}\Mat{Y}^{\frac{1}{2}}\sum_{i=1}^{N}(w_{i}\Mat{L}_{i})\Mat{Y}^{\frac{1}{2}}\Mat{Y}^{-1}\Mat{Y}^{\frac{1}{2}}\sum_{i=1}^{N}w_{i}\Mat{L}_{i})\textrm{ ,}\\
		&=\argmin_{ \{w_i\}_{i=1}^N} \operatorname{Tr}((\sum_{i=1}^{N}w_{i}\Mat{L}_{i})^2)\textrm{ ,} \\
		& s.t.\textrm{  } \sum_{i=1}^N w_i = 1,\, w_i \geq 0\textrm{ ,} 
	\end{split}
\end{tequation}\noindent
where $\operatorname{Tr}$ is the trace of a matrix and  $\Mat{L}_{i}=\operatorname{log}(\Mat{Y}^{-\frac{1}{2}}\Mat{X}_{i}\Mat{Y}^{-\frac{1}{2}})$.
Note that our MCCM-FM formulation is similar to the sparse coding formulation discussed in~\cite{xie2013nonlinear}. 
Different from the work presented in~\cite{xie2013nonlinear}, our work does not consider the sparsity constraint. 
Also, unlike the sparse coding which does not impose positive weights, we constrain the 
weights to be always positive.
In addition, we solve the optimization problem for each class separately.    

\subsection{Manifold Convex Class Model based on a Confined Set — MCCM-CS --- MCCM-CS}\label{SPD-GE}

Solving the optimization problem presented in Eqn.~\eqref{eq:intr_model} can be computationally expensive.
To reduce the computational complexity, 
we propose another approach by relaxing the regularization of Eqn.~\eqref{Eqn:cv_spd_mach1}.
More specifically, we relax the weighted Fr{\'e}cher mean constraint by considering the following property (refer to~\cite{lawson2011monotonic} for proof):

\begin{tequation}
	{\tilde{\Mat{X}} =\argmin_{\tilde{\Mat{X}}} \sum_{i = 1}^N w_i {\operatorname{d_g ^2}}( \Mat{X}_i, \tilde{\Mat{X}})  \textrm{ ,}}
\end{tequation}\noindent
where $\tilde{\Mat{X}} \preceq \sum\limits_{i = 1}^N {{w_i}{\Mat{X}_i}}$. Note that  $\preceq$ represents the Loewner order. $\tilde{\Mat{X}} \preceq \sum\limits_{i = 1}^N {{w_i}{\Mat{X}_i}}$ means $\tilde{\Mat{X}}-\sum\limits_{i = 1}^N {{w_i}{\Mat{X}_i}}$ is positive semi-definite.
Thus, one can utilize $\sum\limits_{i = 1}^N {{w_i}{\Mat{X}_i}}$ as a relaxation of the regularization in Eqn.~\eqref{Eqn:cv_spd_mach1}. 
However, to ensure the distance function:
\begin{tequation} \label{eq:dci}
	\operatorname{d_{cvx}}^2(\Mat{Y}, \mathcal{C}_{SPD}) = \min {\operatorname{d_g ^2}}(\Mat{Y},\sum_{i=1}^N w_i \Mat{X}_i )\textrm{ ,}
\end{tequation}\noindent 
is convex, one needs to further narrow down the set representing the convex model using weights from the set:
{\small
	$\mathcal{A}:= \left\lbrace
	w|\sum\limits_{i = 1}^N {{w_i}{\Mat{X}_i}}  \preceq \Mat{Y},\\
	\sum\limits_{i = 1}^N {{w_i}} = 1\textrm{ ,} {w_i} \ge
	0\right\rbrace\textrm{.}$}
This means $\Mat{Y}-\sum\limits_{i = 1}^N {{w_i}{\Mat{X}_i}} $ is positive semi-definite.
In other words, the convex model can be constructed by only using the weighted convex combinations of samples which under-determine the query point $\Mat{Y}$.
For the convexity proof of the function in Eqn.\eqref{eq:dci}, we refer readers to~\cite{cherian2014riemannian}.
Note that, the difference between our work and the work in~\cite{cherian2014riemannian} 
is that we restrict the sum of $w_i$ to be equal to one and the optimization is based on each class.

Following~\cite{cherian2014riemannian}, to solve the above optimization problem on this restricted set, we first rewrite the objective function in Eqn.~\eqref{eq:dci} by setting {\small$\Mat{M}=\sum\limits_{i = 1}^N {{w_i}{\Mat{X}_i}}$}:

\begin{tequation}
	\begin{split}
		\operatorname{f}(\Vec{w})&={ \operatorname{d_g ^2}}(\Mat{Y},\sum_{i=1}^N w_i \Mat{X}_i )
		=||\operatorname{log}(\Mat{Y}^{- \frac{1}{2}}\Mat{M} \Mat{Y}^{- \frac{1}{2}})||^2_F\\
		&=\operatorname{Tr}\left\lbrace \operatorname{log}(\Mat{Y}^{-\frac{1}{2}}\Mat{M}\Mat{Y}^{- \frac{1}{2}})^\top\operatorname{log}(\Mat{Y}^{-\frac{1}{2}}\Mat{M}\Mat{Y}^{- \frac{1}{2}})\right\rbrace\textrm{ ,}
	\end{split}
\end{tequation}\noindent
where $\Vec{w}$ is an N-dimensional vector whose elements are the set of linear combination weights $\{ w_i \}_{i=1}^N$.
Then the partial derivative of the above function 
can be defined as follows:{
	\begin{tequation}
		\label{d_f}
		\begin{split}
			\frac{\partial\operatorname{f}}{\partial w_i}= 2\operatorname{Tr} \left\{ \operatorname{log}(\Mat{Y}^{-\frac{1}{2}}\Mat{M}\Mat{Y}^{- \frac{1}{2}})(\Mat{Y}^{- \frac{1}{2}}\Mat{M}\Mat{Y}^{- \frac{1}{2}})^{-1}
			\times \frac{\partial( \Mat{Y}^{- \frac{1}{2}}\Mat{M}\Mat{Y}^{- \frac{1}{2}})} {\partial w_i} \right\}\textrm{ ,}
		\end{split}
	\end{tequation}
	\noindent
	where
	\begin{tequation}
		\label{d_m}
		\begin{split}
			{ \frac {\partial(\Mat{Y}^{- \frac{1}{2}}\Mat{M}\Mat{Y}^{- \frac{1}{2}})}{\partial w_i}
				=\Mat{Y}^{- \frac{1}{2}}  \frac{\partial \Mat{M} }{\partial w_i}    \Mat{Y}^{- \frac{1}{2}}
				=\Mat{Y}^{- \frac{1}{2}}\Mat{X_i}\Mat{Y}^{- \frac{1}{2}}\textrm{ .}\nonumber}
		\end{split}
	\end{tequation}\noindent
	Substituting Eqn.$~\eqref{d_m}$ to Eqn.$~\eqref{d_f}$, we have: 
	{\small
		\begin{tequation}
			\frac{\partial\operatorname{f}}{\partial w_i}=2\operatorname{Tr}\left\{\operatorname{log}(\Mat{Y}^{- \frac{1}{2}}\Mat{M}\Mat{Y}^{- \frac{1}{2}})\Mat{Y}^{\frac{1}{2}}\Mat{M}^{-1}\Mat{X}_{i}\Mat{Y}^{-\frac{1}{2}}\right\}\textrm{ .}
	\end{tequation}}
	\noindent
	In the implementation, we used the Spectral Projected Gradient method (SPG)~\cite{schmidt2009optimizing,birgin2001algorithm} to solve the optimization. 
	We name this proposed solution as MCCM-CS.
\subsection{Manifold Convex Model Based on LE Metric--- MCCM-LE}
\label{SPD-LE}
In this section, we describe an approach, here called MCCM-LE, based on the LE Metric presented in Eqn.~\eqref{Eqn:dist_LED}.
This is due to the high computational complexity suffered by the AIRM and one of the advantages of 
using LE metric is that the calculation could be hundreds of times faster.
More importantly, it has been shown that the weighted mean solution calculated using the 
LE metric is similar or in some cases is even equal to the solution of the weighted Fr{\'e}chet mean calculated using the AIRM~\citep{arsigny2007geometric}.
This gives us confidence that the approximation of the convex class model utilizing LE metric will not severely affect the accuracy. 
We can rewrite the optimization problem presented in Eqn.~\eqref{Eqn:cv_spd_mach1} as follows:
{\small
	\begin{tequation}
		\setlength{\abovedisplayskip}{3pt}
		\setlength{\belowdisplayskip}{3pt}
		\label{Eqn:led_convex}
		\begin{split}
			&\operatorname{d_{cvx}}^2(\Mat{Y},  \mathcal{C}_{SPD} )=\min \left\| \operatorname{log}\Mat{Y}- \sum_{i=1}^N  w_i \operatorname{log}\Mat{X}_i \right\|^2_F\textrm{, } \\
			&s.t. \sum_{i=1}^N w_i = 1\textrm{, } w_i \geq 0\textrm{ ,}
		\end{split}
\end{tequation}}\noindent
where $\| \cdot \|_F$ is the Frobenius' norm; $\operatorname{log}( \cdot )$ is the matrix logarithm function that maps from the manifold space into the tangent space at the identity.
Note that as the tangent space at the identity is Euclidean, $\operatorname{log} (\tilde{\Mat{X}})$ can now be represented explicitly as {\small$\operatorname{log} (\tilde{\Mat{X}}) = \sum_{i=1}^N ( w_i \Mat{X}_i)$}.
Expanding the above equation, we can express the optimization problem as:
\begin{small}
	\begin{align}\label{eq:led_convex}
		\setlength{\abovedisplayskip}{2pt}
		\setlength{\belowdisplayskip}{2pt}
		& \argmin_{\{w_i\}_{i=1}^N} \left\| \operatorname{log}\Mat{Y}- \sum_{i=1}^N w_i \operatorname{log}\Mat{X_i} \right\|_F^2 
		=\argmin_{ \{w_i\}_{i=1}^N} [ (\operatorname{log}\Mat{Y}\cdot\operatorname{log}\Mat{Y})\nonumber\\
		& -2\sum\limits_{i = 1}^N {{w_i}(\operatorname{log}\Mat{Y}\cdot\operatorname{log}\Mat{X}_i)} 
		+\sum\limits_{i = 1}^N\sum\limits_{j =1}^N{w_i}{w_j}(\operatorname{log}\Mat{X}_i\cdot\operatorname{log}\Mat{X}_j) ] \textrm{,  }\nonumber \\
		& s.t.\textrm{  }  \sum_{i=1}^N w_i = 1,\, w_i \geq 0\textrm{ . }  
\end{align}\end{small}\noindent
Since Euclidean geometry applies in the space generated by the LE metric,
we can simply vectorize the point {\small$\Mat{A} = \operatorname{log}(\Mat{X})$} as shown in~\citep{pennec2006riemannian}:
{\small$\operatorname{Vec}(\Mat{A})=[a_{1,1}, \sqrt{2}a_{1,2}, a_{2,2},  \sqrt{2}a_{1,3}, \sqrt{2}a_{2,3}, a_{3,3},
	...,\sqrt{2}a_{1,d},...,\sqrt{2}a_{d-1,d}, a_{d,d}]^\top\textrm{. }$}
\noindent
Also, the term
$\operatorname{log}\Mat{Y}\cdot\operatorname{log}\Mat{Y}$ in Eqn.~\eqref{eq:led_convex} is constant and can be excluded from Eqn.~\eqref{eq:led_convex}.
Finally, the optimization problem is rewritten as:
\begin{tequation}
	\setlength{\abovedisplayskip}{3pt}
	\setlength{\belowdisplayskip}{3pt}
	\label{opt_log}
	\begin{split}
		\mathop {\argmin}\limits_{\Vec{w}} \Vec{w}^\top \Mat{D} w-2\operatorname{Vec}({\operatorname{log}}\Mat{Y})\Mat{D}\textrm{ , }
		s.t.\textrm{ }  \Vec{e}^\top \Vec{w}=1,\Vec{w}\ge \Vec{0}\textrm{ , }
	\end{split}
\end{tequation}\noindent
where $\Mat{D}=[\operatorname{Vec}(\operatorname{log}\Mat{X}_1) \cdots \operatorname{Vec}(\operatorname{log}\Mat{X}_N)]$, $\Vec{e}=[1 \cdots 1]^\top$. The above problem is a quadratic optimization problem and we used the public code developed by Cevikalp~\etal ~\cite{cevikalp2010face} for our experiments.

\section{Experimental Results}
\label{ex_results}
To demonstrate the effectiveness of our proposed MCCM framework,
we evaluate the proposed approaches on four computer vision tasks:
object recognition using ETH80 object dataset~\cite{leibe2003analyzing},
texture classification using Brodatz dataset~\cite{randenHusoy1999}, 
person re-identification using ETHZ dataset~\cite{ess2007depth} and
traffic scene classification using UCSD traffic dataset~\cite{chan2005probabilistic}. 
Note that, these four datasets possess limited training data, such as the number of training data is quite small or the data is imbalanced. 
To obtain a reliable performance statistics,
the experiment is repeated ten times. The average performance is then reported.

\begin{figure}[!b]
	\centering
	\includegraphics[width=0.4\textwidth,keepaspectratio]{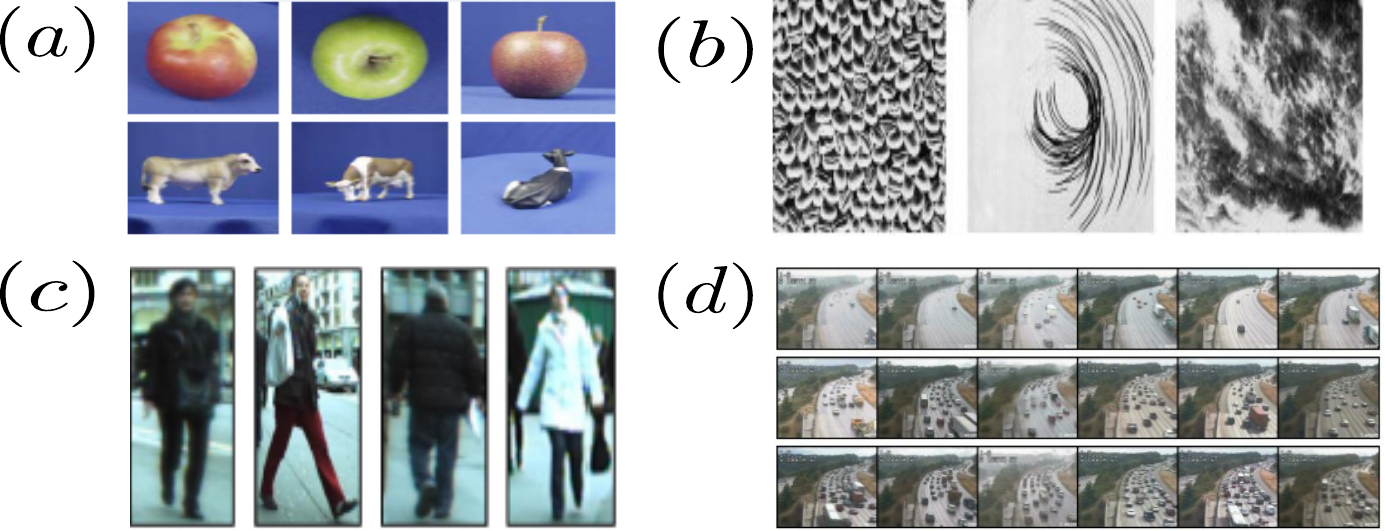}
	\caption{ {\small
			(a) Sample images from ETH-80 object dataset~\cite{leibe2003analyzing};
			(b)~Sample images from BRODATZ texture dataset~\cite{randenHusoy1999};
			(c) Example images from ETHZ data set~\cite{ess2007depth};
			(d) Example frames from UCSD traffic dataset~\cite{chan2005probabilistic}.}}
	\label{fig:dataset}\picspace
\end{figure}                                  
For our comparisons, we considered three baseline classifiers on SPD manifolds: 
(1)~\textbf{Geodesic Nearest Neighbor~(Geo-NN)}--- the geodesic distance in Eqn.~\eqref{Eqn:dist_SPD} was used.
(2)~\textbf{Kernel SVM~(KSVM)}--- an effective implementation of KSVM, LibSVM~\cite{CC01a}, was used in conjunction with the Stein divergence kernel~\cite{sra2011} which has excellent performance for SPD manifolds~\cite{harandisparse}. 
(3)~\textbf{Sparse coding~(SPD-Sc)}--- the state of the art solution of SPD sparse coding proposed in~\cite{cherian2014riemannian} was used. Once the sparse coefficients were determined, the sparse coding classifier proposed in~\cite{wright2009robust} was used as the classifier.

\noindent             
{\bf Object Recognition Using ETH80 Dataset---}
The ETH80 dataset~\cite{leibe2003analyzing} contains eight object categories with ten different object instances in each category.
For each object instance, it provides 41 images of various views as a image set~(refer to Figure~\ref{fig:dataset}~(a) for examples). We use the same protocol for all methods. More specifically, we used five randomly chosen instances per category for training. The other five image sets per category for testing. To generate SPD points,
we used the 80-dimensional DCT coefficients as the feature vector for each image and then computed the $80\times 80$ covariance matrix of each image set.

\begin{table}[!t]
	\centering
	\caption
	{Average accuracy on ETH80 dataset~\cite{leibe2003analyzing}
	}
	\resizebox{0.8\columnwidth}{!}{
		\begin{tabular}{|c|c|}
			\hline
			{ Method}
			&{ Accuracy~(\%)}\\
			\hline
			\hline
			{CHISD~\cite{cevikalp2010face}}&$73.5$\\
			{AHISD~\cite{cevikalp2010face}}&$77.3$\\
			{SANP~\cite{hu2011sparse}}&$75.5$ \\
			{SPD-Sc~\cite{cherian2014riemannian}}&$88.5$ \\
			{KSVM}&$89.5$ \\
			{ Geo-NN}
			&$87.3$ \\
			{MCCM-FM~(proposed)}&$92.3$\\
			{MCCM-CS~(proposed)}
			&${90.5}$ \\
			{MCCM-LE~(proposed)}
			&${\bf93.3}$\\
			\hline
	\end{tabular}}
	\label{tab:eth80}\capspace
\end{table} 

The experimental results for this dataset are summarized in
Table~\ref{tab:eth80}. The maximum accuracy of 93.3\% is achieved
by MCCM-LE, which is nearly six percentage points better than Geo-NN.
We also compare our proposed methods with the Convex model based Image
Set Distance~(CHISD)~\cite{cevikalp2010face},Affine Hull based Image Set Distance~(AHISD) and Sparse
Approximated Nearest Points~(SANP)~\cite{hu2011sparse} methods.
Our MCCM methods outperform all these methods.
Note that CHISD, AHISD and SANP are Euclidean-based convex model methods. 
The performance improvement of the proposed approaches compared to CHISD,AHISD and SANP is significant. 
This could be attributed to using manifold features and the nearest convex class model to perform this classification task.


\noindent {\bf Texture Classification Using Brodatz Dataset---}
We followed the protocol presented in~\cite{sivalingam2010tensor} for this dataset.
This protocol includes three subsets with different numbers of classes: 5-class-texture; 10-class-texture and 16-class-texture.
Each image was resized to $256\times 256$ pixels and divided into 64 regions.
A feature vector for each pixel was calculated using
the grayscale intensity and absolute values of the first- and second-order derivatives of spatial features:
$
F(x, y)
\mbox{~=~}
\left[
I\left(x,y\right),
\left| \frac{\partial I}  {\partial x}  \right|,  \left|\frac{\partial I}  {\partial y}  \right|,
\left| \frac{\partial^2 I}{\partial x^2}\right|,  \left|\frac{\partial^2 I}{\partial y^2}\right|
\right] \textrm{.}
$
Then, each region was represented by a covariance matrix formed from these feature vectors.
For each scenario, five SPD points per class were randomly selected as training, and the rest were used for testing.
\begin{table}[!t]
	\centering
	\capspace
	\caption
	{Average accuracy on Brodatz dataset~\cite{randenHusoy1999}}
	\label{tab:BRO}
	\resizebox{0.8\columnwidth}{!}{
		\begin{tabular}{|c|c|}
			\hline
			{ Method}
			&{ Accuracy~(\%)}\\
			\hline
			\hline
			~{ LE-SR~\cite{SR_Riemannian_AVSS_2010}}&$66.3$\\
			{ TSC~\cite{sivalingam2010tensor}}&~$79.8$\\
			{ RLPP~\cite{harandikernel}}~&$86.1$\\
			{SPD-Sc~\cite{cherian2014riemannian}}&$77.6$\\
			{ KSVM}&$82.6$\\
			{ Geo-NN}&$84.9$\\ 
			{MCCM-FM~(proposed)}&$\bf{88.0}$\\
			{ MCCM-CS~(proposed)}&$\bf{88.0}$\\
			{ MCCM-LE~(proposed)}&$87.8$\\
			\hline
	\end{tabular}}
\end{table} 

Table~\ref{tab:BRO} compares the proposed MCCM-FM,  MCCM-CS and MCCM-LE
to various methods. 
Note that the number of training data per class is only five.
The proposed methods significantly outperform all the baselines.
This corroborates the previous findings in the Euclidean space~\cite{zhou2009nearest},
that suggest the manifold convex class model is generally more robust to a small number of training samples.
Compared to the recent methods such as Log-Euclidean Sparse Representation~(LE-SR)~\cite{SR_Riemannian_AVSS_2010},
Tensor Sparse Coding~(TSC)~\cite{sivalingam2010tensor} and Riemannian Locality Preserving
Projection~(RLPP)~\cite{harandikernel}, the 
proposed methods perform better.

\begin{table}[!t]
	
	\centering
	\capspace
	\caption
	{ Average accuracy on the ETHZ dataset~\cite{ess2007depth}
	} \label{tab:ETHZdataset}
	\resizebox{0.8\columnwidth}{!}{
		\begin{tabular}{|c|c|}
			\hline
			{Method}
			&{ Accuracy (\%)}\\
			\hline
			\hline
			{ PLS~\cite{schwartz2009learning}}&$77.3$ \\
			{ HPE~\cite{bazzani2010multiple}}&$84.2$ \\
			
			{ SPD-Sc~\cite{cherian2014riemannian}}&$90.1$ \\
			{ KSVM}&$89.5$\\
			{ Geo-NN}&$87.1$\\ 
			{MCCM-FM~(proposed)}&$\bf{93.4}$\\
			{ MCCM-CS~(proposed)}&${92.0}$\\
			{MCCM-LE~(proposed)}&$90.6$\\
			\hline
	\end{tabular}}
\end{table}

\noindent
{\bf Person Re-identification Using ETHZ Dataset---}
The ETHZ dataset~\cite{ess2007depth} was captured from a moving camera, containing wide variations in the appearance of people~(refer to Figure~\ref{fig:dataset}~(c) for examples).
The dataset is divided into three sequences.
Sequence 1 contains 83 pedestrians captured in 4,857 images, 
sequence 2 contains 35 pedestrians captured in 1,936 images, and sequence
3 contains 28 pedestrians captured in 1,762 images.
In this experiment,  to demonstrate the advantage of our proposed methods, only 10 randomly selected images per class were used as training set, while the rest were formed as testing.
To generate the SPD features, we first resized all the images into $64\times 32$ pixels.
Then the SPD features were generated by computing covariance matrix of the pixels feature vectors defined as:
\begin{equation}
	F_{x,y}\mbox{=}
	\left[
	x, y,
	R_{x,y},   G_{x,y},   B_{x,y},
	R_{x,y}',  G_{x,y}',  B_{x,y}',
	R_{x,y}'', G_{x,y}'', B_{x,y}''
	\right]\textrm{, }\nonumber
\end{equation}
\noindent where {\small $x$} and {\small $y$} represent the position of a pixel, while
{\footnotesize $R_{x,y}$}, {\footnotesize $G_{x,y}$} and {\footnotesize $B_{x,y}$}
represent the corresponding color information, respectively.
In this experiment, 10 randomly selected images per class were used as training set, while the rest were formed as testing.


Table~\ref{tab:ETHZdataset} shows the performance of different methods on this dataset. The proposed methods are the best two
among all methods and achieve considerably better results than
the three baselines:~Geo-NN, KSVM and SPD-Sc. 
Furthermore, 
the proposed methods significantly outperform
Partial Least Squares~(PLS) \cite{schwartz2009learning} and Histogram Plus Epitome~(HPE)~\cite{bazzani2010multiple}.
This could be attributed to the efficacy of our proposed classifier with manifold features.

\begin{table}[!t]
	\centering 
	\caption
	{
		Average accuracy on UCSD dataset~\cite{chan2005probabilistic}
	}
	\resizebox{0.8\columnwidth}{!}{
		\begin{tabular}{|c|c|}
			\hline
			{ Method}
			&{ Accuracy~(\%)}\\
			\hline
			\hline
			{ LDS~\cite{san2010compressive} } 
			&$87.5$ \\
			{ CS-LDS~\cite{san2010compressive}}&$89.1$  \\
			{SOA~\cite{derpanis2011classification}}&$\bf 95.2$\\
			{  DNLSSA - RBF Kernel~\cite{baktashmotlagh2014discriminative}}&$94.5 $  \\
			{SPD-Sc~\cite{cherian2014riemannian}}&$90.9$ \\
			{KSVM}&$93.7$ \\
			{ Geo-NN} 
			&$91.3$  \\
			{MCCM-FM~(proposed)}&${94.1}$\\
			{MCCM-CS~(proposed)} 
			& ${ 94.5}$  \\
			{ MCCM-LE~(proposed)} 
			&$94.1$ \\
			
			\hline
		\end{tabular}
	}
	\label{tab:UCSD}
\end{table}

\noindent{\bf Traffic Scene Classification Using UCSD Dataset---}
This dataset comprises 254 video sequences
collected from the highway traffic in Seattle over two days~\cite{chan2005probabilistic}~(see Figure~\ref{fig:dataset}~(d) for examples).
There are three different classes: heavy traffic~(44 sequences),
medium traffic~(45 sequences) and light traffic~(165 sequences).
This dataset is unbalanced since the light traffic data is nearly four times the number of heavy traffic data. 
We follow the common practice~\cite{chan2005probabilistic} to split this dataset into the training and test. 
To generate the SPD points, each frame in one sequence was downsized to $140\times161$ pixels and further normalized by subtracting the mean frame and dividing the variance.
Then, we applied the two dimensional Discrete Cosine Transform (DCT) on the frame and used the DCT coefficients as the feature vector for each frame.
In the consideration of successive frame variation of each sequence, we generated the SPD manifold features by computing the covariance matrix of 15 successive frames in each video sequence.

We compare the performance of our convex class model methods, MCCM-FM,
MCCM-CS and MCCM-LE with the baselines, Geo-NN, KSVM and SPD-Sc.
Table~\ref{tab:UCSD} shows our methods outperform all of these baselines.
In addition, Table~\ref{tab:UCSD} also presents the performance of the recent methods such as Linear Dynamical Systems
model~(LDS)~\cite{san2010compressive}, Compressive Sensing
LDS~(CSLDS)~\cite{san2010compressive}, Spatio-temporal
Orientation Analysis (SOA)~\cite{derpanis2011classification} and Non-Linear Stationary
Subspace Analysis~(DNLSSA)~\cite{baktashmotlagh2014discriminative}. 
The accuracy of our methods is considerably better than both LDS and CSLDS.
We achieve competitive performance to DNLSSA and SOA.
It is noteworthy to mention that DNLSSA and SOA are considerably more complex.
As SOA requires matching distributions of space-time orientation structure
and DNLSSA solves an optimization problem in the kernel space with manifold regularization. 
Furthermore, our methods do not require any parameter tuning.

It worthy to note that our proposed MCCM methods achieved much better performance than SPD-Sc in all experiments, especially on Brodatz~($10.4$ percentage points better). We conjecture that the improvement could be attributed to the regularization used on the weights $w_i$ as the convex model constraints and modeling each individual class as one convex model.

\section{Further Analysis}
\label{further}
In this paper, we have discussed our experiment results suggesting that our proposed methods outperform the baselines and the recent methods in each dataset.
In this part we perform further study on the three proposed methods.
In particular, two analysis are presented: (1) running time analysis; (2) analysis on the approximation error generated by the proposed method and
(3) comparisons with Geo-NN in terms of sensitivity to noisy data and limited training data.

\subsection{Running time analysis}
To perform this analysis, we timed each method running time to perform the whole experiment comprising calculating the prediction label for each query. 
Note that our approach does not need time for training.  The experiments were performed on an Intel 3.40 GHz processor using Matlab.
{\small \begin{table}[!b]
		\centering
		\resizebox{0.85\columnwidth}{!}{
			\begin{tabular}{|c|c|c|c|c|}
				\hline
				{ Method}&{ ETH80}&{ BRODATZ}&{ ETHZ}&{ UCSD}\\
				\hline
				\hline
				{MCCM-FM}&$46.04s$&$279.62s$&$28486.08s$&$110.03s$\\
				{ MCCM-CS}&$46.15s$&$160.74s$&$2338.49s$&$19.83s$\\
				{ MCCM-LE}&${\bf 3.74s}$&${\bf33.09s}$&${\bf1511.71s}$&${\bf2.64s}$\\
				
				\hline
				
		\end{tabular}}
		\vspace{0.7ex}
		\caption
		{
			Time comparison: The run time of MCCM-FM, MCCM-CS and MCCM-LE on each dataset. 
		}
		\label{tab:time}

\end{table} } 

Table~\ref{tab:time} shows the running time for the three methods on the four applications. 
As MCCM-FM requires optimization process based on Fr{\'e}chet mean~(See Eqn.(15) Section 4.2), it becomes the slowest method of the three. 
MCCM-CS using the matrix additions which will speed up the computation, 
especially on  ETHZ dataset~(12.2 times faster than MCCM-FM) and UCSD dataset~(5.5 times faster than MCCM-FM).
Compared to MCCM-FM, MCCM-LE is significantly more efficient.
The speed up achieved by MCCM-LE is 41.7 times on UCSD dataset, 18.8 times on ETHZ dataset, 12.3 times on ETH80 dataset and 8.5 times on BRODATZ dataset.
It is noteworthy that despite its extremely fast running time, MCCM-LE does not suffer much performance loss.
The worst drop in accuracy of MCCM-LE is reported in ETHZ dataset with 2.8 percentage point drops from MCCM-FM.
However, given the 18.8 times speed up gained by MCCM-LE, this performance drop is an excellent trade off.
This also corroborates the work of Arsigny~\etal~\cite{arsigny2007geometric} stating that the solution of the weighted mean calculated using LE metric is fairly similar or in some cases could be the same.
All our methods depend on the weighted mean accuracy to compute the convex model distance.

\subsection{Analysis of approximation errors}
The crux in solving the optimization problem to calculate the convex model distance in Eqn.(13)~(refer to Section 4.2) is on the implicit formulation of the weighted mean.
As shown in MCCM-FM, MCCM-CS and MCCM-LE, once this is formulated in an explicit form, the problem can be addressed.
To achieve this, we perform three variants: 
\begin{itemize}
	\item
	MCCM-FM exploits the tangent space at the query point. In this space, any distance to the query point represents the true manifold geodesic distance. Thus, this allows us to consider the manifold geometric structure whilst solving the optimization problem. However, this has an assumption that requires the query and the convex model should be relatively close.
	\item
	MCCM-CS considers Euclidean linear combination. The set of points produced by this form is said to be under-determined by the  mean~\cite{lawson2011monotonic}, which means instead of seeking an arbitrary approximation of the weighted Fr{\'e}chet mean points, we confine all the approximate points to have the Loewner partial order with the Fr{\'e}chet mean.
	\item
	MCCM-LE is motivated by the work of Arsigny~\etal\cite{arsigny2007geometric} suggesting that the Fr{\'e}chet  mean solution using the LE metric is similar or the same as the Fr{\'e}chet mean solution using the AIRM.
\end{itemize}

\begin{figure}[!b]
	\centering
	\includegraphics[width=0.4\textwidth,keepaspectratio]{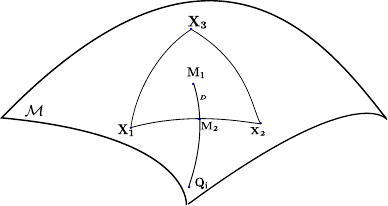}
	\caption{Experiment set-up for the approximation error analysis. $\Mat{X_1}$, $\Mat{X_2}$, $\Mat{X_3}$ are randomly generated SPD points; $\Mat{M_1}$ is the Fr{\'e}chet  mean of these three points; $\Mat{M}_2$ is the Fr{\'e}chet  mean of $\Mat{X}_1$ and $\Mat{X}_2$; $D$ denotes the geodesic distance from $\Mat{M}_1$ and $\Mat{M}_2$.}
	\label{fig:ex_set}
	
\end{figure}
Given these approximations, one may ask how much impact of 
the errors generated by these in calculating the convex model distance is.
To answer this, we perform an experiment using a synthetic dataset.
In this experiment, a query, $\Mat{Y}$ and a convex model, $\mathcal{C}_{SPD}$ were given. 
We assume that the true $\tilde{\Mat{X}}$ which minimizes Eqn.(13) (refer to Section 4.2) 
was also given.
We then used MCCM-FM, MCCM-CS and MCCM-LE to compute the convex model distance from $\mathcal{C}_{SPD}$ to $\Mat{Y}$.
Unequivocally, as we already knew the true $\tilde{\Mat{X}}$, then it was easy to compute the correct convex model distance.
Finally, we evaluated the error by the absolute value of the difference between the correct distance and the estimated distance.
In addition, we evaluated the error when the query point was progressively moved further away from the convex model.

To simplify the analysis, we considered a three-point convex model.
Note that the results from this analysis could be easily extrapolated to $n$-point convex models.
Here, we constructed a convex model $\mathcal{C}_3$ by using three randomly generated SPD points $\Mat{X}_1$, $\Mat{X}_2$ and $\Mat{X}_3$.
Let $\Mat{M}_1$ be the mean of these points.
The three points were generated in such a way that the geodesic distance between every point to $\Mat{M}_1$ is equal.

Let $\Mat{M}_2$ be the mean of $\Mat{X}_1$ and $\Mat{X}_2$ and the geodesic distance from $\Mat{M}_1$ and $\Mat{M}_2$ as $D$.
We generated a set of query points $\{ \Mat{Q}_{i} \}_{i=1}^4$, along the geodesic from $\Mat{M}_1$ to $\Mat{M}_2$.
By doing this, the correct nearest point from $\Mat{Q}_i$ to the convex model $\mathcal{C}_3$ is $\Mat{M}_2$ (\ie $\tilde{\Mat{X}} = \Mat{M}_2$).
This is because, $\Mat{Q}_i$ is on the geodesic passing $\Mat{M}_2$ and $ \Mat{M}_1 \in \mathcal{C}_3$ and there can be only one unique geodesic in SPD manifolds~\cite{pennec2006riemannian}.

As mentioned, we progressively moved the query point $\Mat{Q}_i$ along the geodesic passing $\Mat{M}_2$ and $\Mat{M}_1$.
The distance from the query points $\{ \Mat{Q}_1 \cdots \Mat{Q}_4 \}$ to $\Mat{M}_1$ was set to $5\times D$, $10\times D$, $100\times D$ and $200\times D$.
Figure~\ref{fig:ex_set} illustrates our experiment set-up.
The tests were repeated 50 times and we reported the average absolute error in Table~\ref{tab:ERROR}.

As expected, the further the query from the convex model, the more error will be produced.
However, we found that the errors of all three solutions were significantly small even when the query point was very far from the convex model.
The error produced by MCCM-FM is the lowest for all settings. MCCM-FM is designed to solve the problem by attempting to preserve the manifold topological structure as much as possible.
MCCM-LE performs as the second best approximation method. 
The error generated by MCCM-CS is higher than MCCM-FM and MCCM-LE, however it is still extremely small.
{\small\begin{table}[!t]
		\centering
		\resizebox{0.8\columnwidth}{!}{
			\begin{tabular}{|c|c|c|c|c|}
				\hline
				{ Distance}&{ $5\times D$}&{ $10\times D$}&{ $100\times D$}&{ $200\times D$}\\
				\hline
				\hline
				{MCCM-FM}&\bf{1.97e-04} &\bf{3.18e-04} &\bf{1.53e-03}
				&\bf{3.35e-03}\\
				{ MCCM-CS}&8.71e-02&8.72e-02&8.72e-02&8.75e-02\\
				{ MCCM-LE}&6.06e-02&6.06e-02&6.14e-02&6.19e-02\\
				
				\hline
		\end{tabular}}
		\vspace{0.7ex}
		\caption
		{
			The average error of MCCM-FM, MCCM-CS and MCCM-LE evaluated using synthetic data.  
		}
		\label{tab:ERROR}
		
\end{table} }
These results provide two suggestions: (1) the approximation used in MCCM-FM is the most accurate as it considers to preserve the manifold structure; (2) the error of all three methods is extremely small. This means, the approximation will not significantly give adverse effect on the accuracy of the classifier utilizing the convex class model. 
This may explain the good performance achieved by all three methods in four datasets.

\subsection{Comparisons with Geo-NN }         
As mentioned in the section~\ref{intro}, Geo-NN is sensitive given a  small number of training data. Our methods tackle this problem by representing each class by a convex model that implicitly increases the number of training data.
In this section, we further demonstrate our advantage over Geo-NN on a more challenging dataset-- Maryland"in-the-wild" dataset~(ML) with thirteen different classes of dynamic scenes~\cite{shroff2010moving}. This dataset contains 130 video samples that capture large variations in illumination, frame rate, viewpoint, image scale and various degrees of camera-induced motion and scene cuts (refer Figure~\ref{fig:ml} for examples). We used the last layer of the CNN trained in~\cite{zhou2014learning} as frame descriptors. We then used PCA to reduce the dimensionality of the CNN features to 400. To generate the SPD point, we computed the covariance matrix of the frame descriptors in one video. Note that, to demonstrate the advantage of our methods for small training sets, we randomly selected five SPD points per class as the training set, the remaining points as the testing set. The experiments were conducted 10 times and the average performance is shown in Figure~\ref{fig:noise}. The performance of our methods are significantly better than Geo-NN.  MCCM-FM achieved 63.7\%, which is 17.7\% better than Geo-NN.
		
In the meantime, we performed an empirical experiment on ETH80 by adding artificial points into the training data for Geo-NN. These artificial points were randomly generated using the weighted Fr{\'e}chet means of the training points~\cite{moakher2005differential}. 
More specifically, a set of random weights were progressively generated for computing the weighted Fr{\'e}chet means of each class.
The resulting means were then added into the training class.
Shown in Figure~\ref{fig:compares}, we found that $160$ artificial training points (four times as large as the number of original training data) was required for Geo-NN to achieve the comparable accuracy with MCCM-FM.
\begin{figure}[!t]
	\centering
	\includegraphics[width=0.4\textwidth,keepaspectratio]{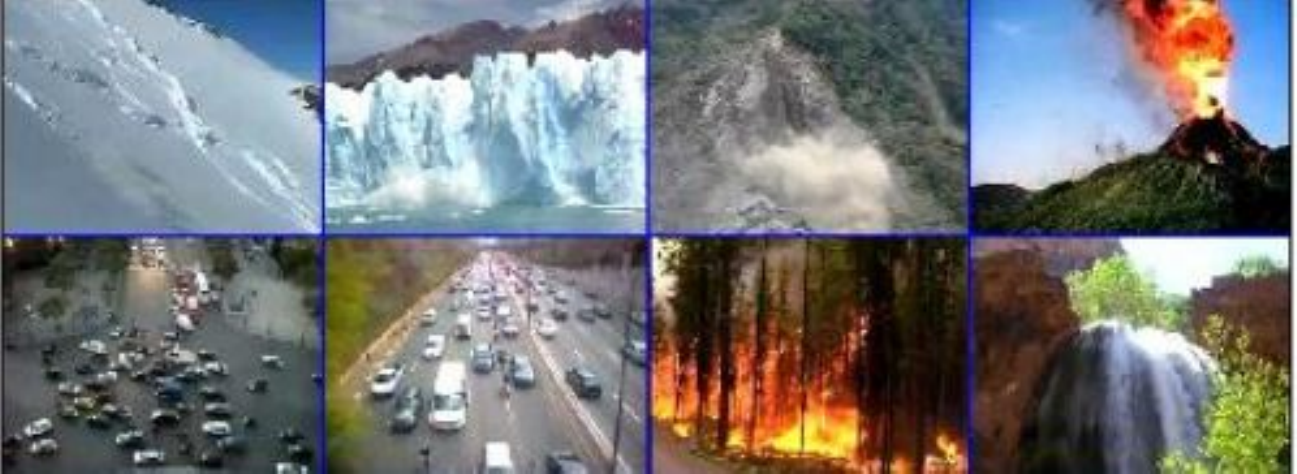}\capspace
	\caption{Examples of Maryland “in-the-wild” scenes data set~\cite{shroff2010moving}. From
		left to right and top to bottom: Avalanche, Iceberg Collapse, Landslide, Volcano eruption, Chaotic traffic, Smooth traffic, Forest fire and Waterfall. }
	\label{fig:ml}
\end{figure}
\begin{figure}
	\centering
	\includegraphics[width=0.5\textwidth,keepaspectratio]{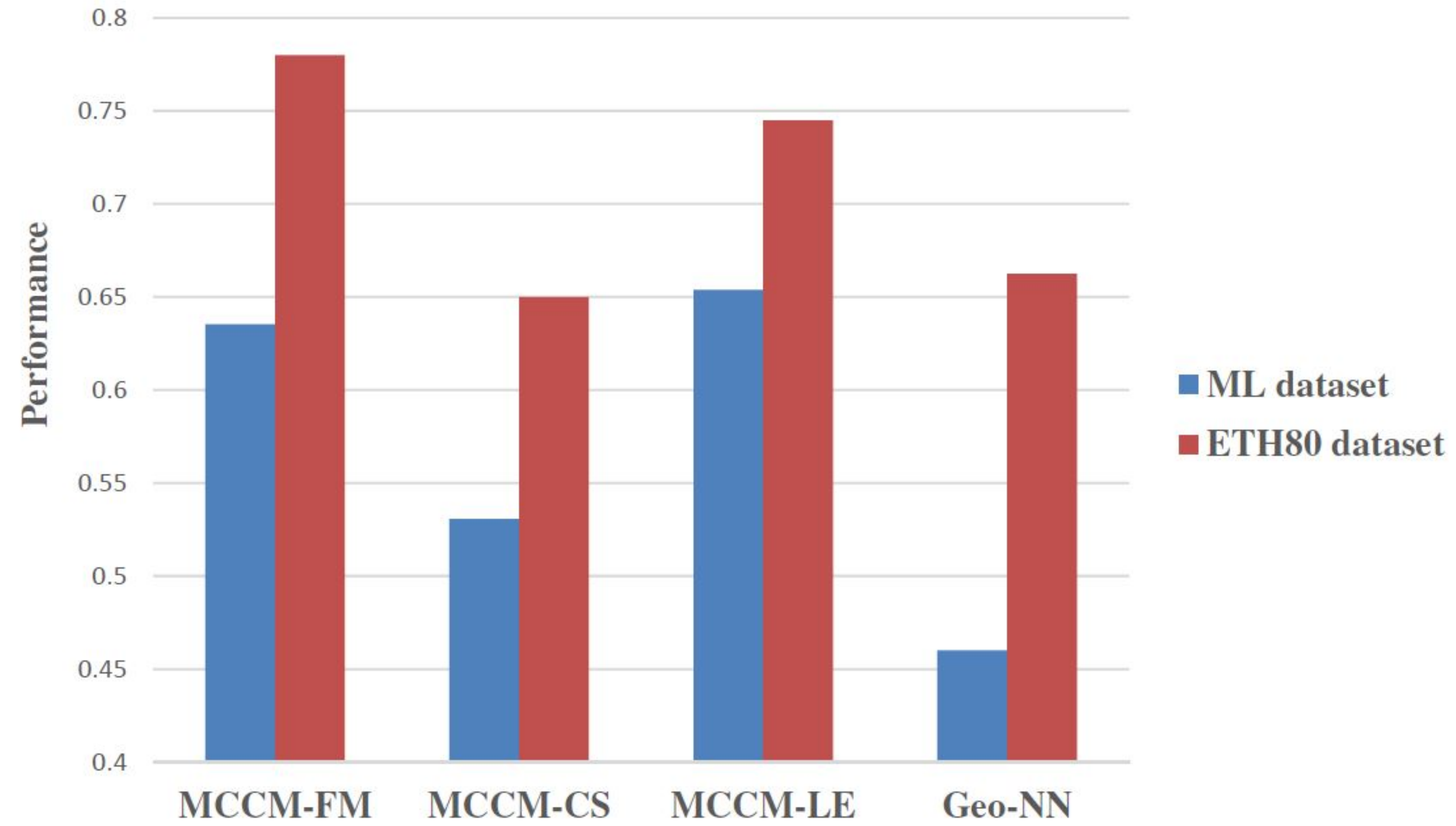}
	\caption{ Comparison results between the proposed methods and Geo-NN on ETH80 dataset with strong
		noise and Maryland~(ML) Dataset. Our proposed methods MCCM performed considerably better than Geo-NN.}
	\label{fig:noise}
	\end{figure}
\begin{figure}[!t]
	\centering
	\includegraphics[width=0.5\textwidth,keepaspectratio]{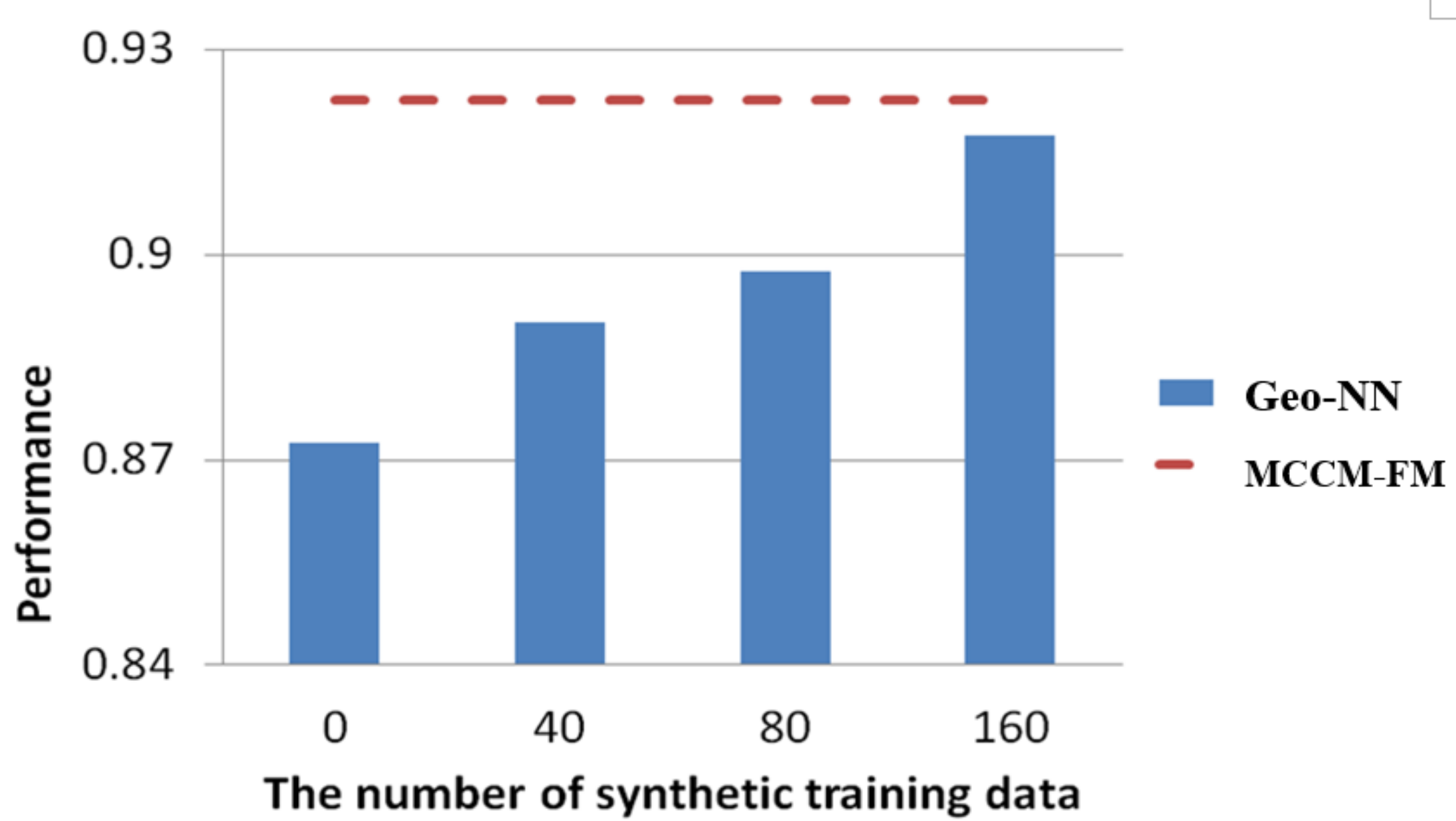}\capspace
	\caption{ Generate synthetic training data for Geo-NN on ETH80 dataset. 160 synthetic points are required by Geo-NN to achieve comparable performance to MCCM-FM. Note that the performance of MCCM-FM is obtained without using any synthetic points. }
	\label{fig:compares}
\end{figure}

Unfortunately, this process is extremely expensive, as the generation of these 160 artificial points which used weighted Fr{\'e}chet mean required 22min.
While, the average running time for all the MCCM methods to perform the whole experiment on the ETH80 is less than a minute.
This indicates the additional advantage of MCCM over the intrinsic approaches such as Geo-NN when only a small number of training data is available.

We also investigated the noise sensitivities of our proposed methods and Geo-NN. In this part of experiments, we added Gaussian noise with zero mean and 0.01 variance to the images of ETH-80 dataset. 
Then we re-generated the SPD features for the training and test dataset using the same process in Section~\ref{ex_results}.    
Results in Figure~\ref{fig:noise} indicate that when strong noises occur in the images, the proposed MCCM-FM method considerably outperforms Geo-NN.  
This suggests that the manifold convex model distance is a much better measurement than Geo-NN for classifying noisy images.  
MCCM-LE is the second best method which is 3.5\% worse than MCCM-FM. MCCM-CS is the one that was most adversely effected by image noise.
We can conclude that, among the different solutions in the convex model framework, MCCM-FM that mostly preserved the intrinsic manifold structure is the best choice when dealing with noisy images.
\section{Conclusions}
\label{conclusion} In this work, we presented the Manifold Convex Class Model~(MCCM) for classification tasks with limited training data on SPD manifolds.
To solve the optimization problem posed when performing nearest convex model computation, we studied three different solutions MCCM-FM, MCCM-CS and MCCM-LE.
MCCM-FM is designed to solve the problem by trying to preserve the manifold structure as much as possible, whilst MCCM-CS and MCCM-LE are the approximate solutions that possess lower computational load.    
Experiments were performed on four computer vision
applications where our proposed methods showed superior performance than several recent methods.
A promising future direction is to utilize the existing deep learning architectures with the objective function designed by our proposed framework.

\section*{Acknowledgements}
This work has been partly funded by Sullivan Nicolaides Pathology, Australia and the Australian Research Council (ARC) Linkage Projects [Grant numbers LP130100230, LP160101797]. Arnold Wiliem is funded by the Advance Queensland Early-Career Research Fellowship.

{ { 
\bibliography{refs1}}
}

\end{document}